\documentclass[11pt]{article}
\usepackage{semdial2018}
\usepackage{times}
\usepackage{latexsym}
\usepackage{graphicx}
\usepackage{amsmath}
\usepackage{lingmacros}
\usepackage{makecell}
\usepackage{xcolor,colortbl}
\usepackage[titletoc,title]{appendix}
\usepackage{caption}
\usepackage{url}
\usepackage{hyperref}

\definecolor{Gray}{gray}{0.90}
\definecolor{LightCyan}{rgb}{0.88,1,1}

\newcolumntype{a}{>{\columncolor{Gray}}c}
\newcolumntype{b}{>{\columncolor{white}}c}

\title{Multi-Task Learning for Domain-General Spoken Disfluency Detection in Dialogue Systems}

\author{Igor Shalyminov, Arash Eshghi, and Oliver Lemon \\
  The Interaction Lab, Department of Computer Science\\
  Heriot-Watt University, 
  Edinburgh, EH14 4AS, UK\\
  {\tt \{is33, a.eshghi, o.lemon\}@hw.ac.uk}\\
  }

\date{}

\begin{document}
\maketitle
\begin{abstract}
  Spontaneous spoken dialogue is often disfluent, containing pauses, hesitations, self-corrections and false starts. Processing such phenomena is essential in understanding a speaker's intended meaning and controlling the flow of the conversation. Furthermore, this processing needs to be \emph{word-by-word incremental} to allow further downstream processing to begin as early as possible in order to handle real spontaneous human conversational behaviour.
  In addition, from a developer's point of view, it is highly desirable to be able to develop systems which can be trained from `clean' examples while also able to generalise to the very diverse disfluent variations on the same data -- thereby enhancing both data-efficiency and robustness. In this paper, we present a multi-task LSTM-based model for incremental detection of disfluency structure\footnote{Code and trained models available at \url{https://bit.ly/multitask_disfluency}},  which can be hooked up to any component for incremental interpretation (e.g. an incremental semantic parser), or else simply used to `clean up' the current utterance as it is being produced.   
  We train the system on the Switchboard Dialogue Acts (SWDA) corpus and present its accuracy on this dataset. Our model outperforms prior neural network-based incremental approaches by about 10 percentage points on SWDA while employing a simpler architecture. To test the model's generalisation potential, we evaluate the same model on the bAbI+ dataset, without any additional training. bAbI+ is a dataset of synthesised goal-oriented dialogues where we control the distribution of disfluencies and their types. This shows that our approach has good generalisation potential, and sheds more light on which types of disfluency might be  amenable to domain-general processing.
\end{abstract}

\section{Introduction}
\label{sec:introduction}
It is uncontested that humans process (parse and generate) language, \emph{incrementally}, word by word, rather than turn by turn, or sentence by sentence \cite{Howes.etal10,Crocker.etal00,Ferreira.etal04}. This leads to many characteristic phenomena in spontaneous dialogue that are difficult to capture in traditional linguistic approaches and are still largely ignored by dialogue system developers. These include various kinds of context-dependent fragment \cite{Fernandez.Ginzburg02b,Fernandez06,Kempson.etal17}, false starts, suggested add-ons, barge-ins and disfluencies.

In this paper, we focus on disfluencies: pauses, hesitations, false starts and self-corrections that are common in natural spoken dialogue. These proceed according to a well-established general structure with three phases \cite{Shriberg94}:

\begin{center}
\begin{footnotesize}
\enumsentence{with $\underbrace{[\textrm{Italian}}_{reparandum}+\underbrace{\{uh\}}_{interregnum} \underbrace{\textrm{Spanish}]}_{repair}$ \ \ cuisine}\label{repair}
\end{footnotesize}
\end{center}

Specific disfluency structures have been shown to serve different purposes for both the speaker \& the hearer (see e.g \newcite{Brennan.Schober01}), for example, a filled pause such as `uhm' can elicit a completion from the interlocutor, but also serve as a turn-holding device; mid-sentence self-corrections are utilised to deal with the speaker's own error as  early as possible, thus minimising effort.

In dialogue systems, the detection, processing \& integration of disfluency structure is thus crucial to understanding the interlocutor's intended meaning (i.e. robust Natural Language Understanding), but also for coordinating the flow of the interaction. 
Like dialogue processing in general, the detection \& integration of disfluencies needs to be \emph{strongly incremental}: it needs to proceed word by word, enabling downstream processing to begin as early as possible, leading to more efficient and more naturally interactive dialogue systems \cite{Skantze.Hjalmarsson10,Schlangen.Skantze09}. 

Furthermore, incremental disfluency detection needs to proceed with minimal latency \& commit to hypotheses as early as possible in order to avoid `jittering' in the output and having to undo the downstream processes started based on erroneous hypotheses \cite{Schlangen.Skantze09,DBLP:conf/emnlp/HoughP14,DBLP:conf/interspeech/HoughS15} .

While many current data-driven dialogue systems tend to be trained end-to-end on natural data, they don't normally take the existence of disfluencies into account. Recent experiments 
have shown that end-to-end dialogue models such as Memory Networks (MemN2N) \cite{babi} 
need impractically large amounts of training data containing disfluencies and with sufficient variation in order to obtain reasonable performance \cite{DBLP:conf/emnlp/EshghiSL17,Shalyminov.etal17}. The problem is that, taken together with the particular syntactic and semantic contexts in which they occur, disfluencies are very sparsely distributed, which leads to a large mismatch between the training data and actual real-world spontaneous user input to a deployed system. This suggests a more modular, pipelined approach, where disfluencies are detected and processed by a separate, domain-general module, and only then any resulting representations are passed on for downstream processing. The upshot of such a modular approach would be a major advantage in generality, robustness, and data-efficiency.

In this paper, we build on the state-of-the-art neural models of \newcite{DBLP:conf/interspeech/HoughS15} and \newcite{DBLP:conf/eacl/SchlangenH17}. 
Our contributions are that: (1) we produce a new, multi-task LSTM-based model with a simpler architecture for incremental disfluency detection, with significantly improved  performance on the SWDA, a disfluency-tagged corpus of open-domain conversations; and (2) we perform a generalisation experiment measuring how well the models perform on unseen data using the controlled environment of bAbI+ \cite{DBLP:conf/emnlp/EshghiSL17}, a synthetic dataset of goal-oriented dialogues in a restaurant search domain augmented with spoken disfluencies.




\section{Related work}

Work on disfluency detection has a long history, going back to \newcite{Charniak.Johnson01} who set the challenge. One of the important dividing lines through this work is the \emph{incrementality} aspect, i.e.\ whether disfluency structure is predicted word by word.

In the non-incremental setting, as the problem is essentially sequence tagging, neural models have been widely used. As such, there are approaches using an encoder-decoder model (seq2seq) with attention \cite{DBLP:conf/coling/WangCL16} and a Stack-LSTM model working as a buffer of a transition-based parser \cite{DBLP:conf/coling/WangCL16,DBLP:conf/emnlp/WangCZZL17}, the latter being state-of-the-art for the non-incremental setting.

Incremental, online processing of disfluencies is a more challenging task, if only because there is much less information available for tagging, viz. only the context on the left. In a practical system, it also involves extra constraints and evaluation criteria such as minimal latency and revisions to past hypotheses which lead to `jittering' in the output with all the dependent downstream processes having to be undone, thus impeding efficiency (see the illuminating discussions in \newcite{DBLP:conf/emnlp/HoughP14} and \newcite{Purver.etal18}). 

Incremental disfluency detection models include  \newcite{DBLP:conf/emnlp/HoughP14} who approach the problem information-theoretically, using local surprisal/entropy measures and a pipeline of classifiers for recognition of the various components of disfluency structure. While the model is very effective, it leaves one desiring a simpler alternative. This was made possible after the overall success of RNN-based models, which \newcite{DBLP:conf/interspeech/HoughS15} exploit. We build on this model here, as well as evaluate it further (see below). On the other hand, \newcite{DBLP:conf/eacl/SchlangenH17} tackle the task of joint disfluency prediction and utterance segmentation, and demonstrate that the two tasks interact and thus are better approached jointly.

Language models have been extensively used for improving neural models' performance. For example, \newcite{DBLP:conf/naacl/PetersNIGCLZ18} showed that a pre-trained language model improves RNN-based models' performance in a number of NLP tasks~--- either as the main feature representation for the downstream model, or as additional information in the form of a latent vector in the intermediate layers of complex models. The latter way was also employed by 
\newcite{DBLP:conf/acl/PetersABP17} in the task of sequence labeling.

Finally, a multitask setup with language modelling as the second objective -- the closest to our approach -- was used by \newcite{DBLP:conf/acl/Rei17} to improve the performance of RNN-based Name Entity Recognition.

We note that there is no previous approach to multitask disfluency detection using a secondary task as general and versatile as language modelling. Furthermore, none of the works mentioned study how well their models {\it generalise} across datasets, nor do they shed much light on what kinds of disfluency structure are harder to detect, and why, as we try to do below.


\section{Disfluency detection model}\label{model}
Our approach to disfluency detection is a sequence tagging model which makes single-word predictions given context words $w_{t-n+1}, ..., w_{t}$ of a maximum length $n$. We train it to perform two tasks jointly (c.f.\ \newcite{DBLP:conf/interspeech/HoughS15}): (1) predicting the disfluency tag of the current word, $P(y_t|w_{t-n+1}, ..., w_{t})$; and (2) predicting the next word in the sequence in a language model way, $P(w_{t+1}|w_{t-n+1}, ..., w_{t})$.

At training time, we optimise the two tasks jointly, but at test time we only look at the resulting tags and ignore the LM predictions.

\begin{figure}
\begin{center}
\includegraphics[width=10cm]{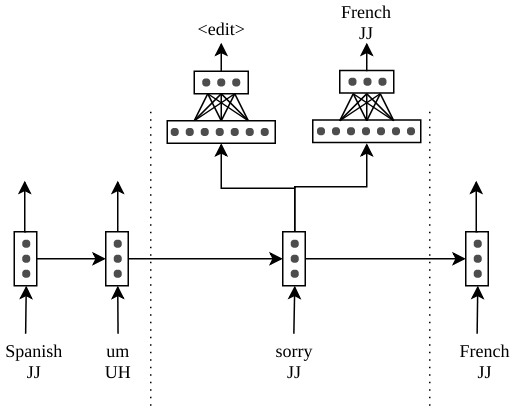}
\end{center}
\caption{Multi-task LSTM model architecture}
\label{fig:model_architecture}
\end{figure}
Our model uses a shared LSTM encoder  \cite{lstm} with combined \texttt{word/POS-tag} tokens which provides context embedding for two independent multilayer perceptrons (MLPs) making the predictions for the two tasks. The combined token vocabulary (word+POS) size for the SWDA dataset is approximately 30\%~ larger than the original word-only version ~--- given this, concatenation is the simplest and most efficient way to pass part-of-speech information into the model.

The intuition behind adding an additional task to optimise for is that it \textit{serves as a natural regulariser}: given an imbalanced label distribution (see Section \ref{sec:data} for the dataset description), only learning disfluency labels may lead to a higher degree of overfitting, and introducing an additional task with more uniformly distributed labels can help the model generalise better.

Other potential benefits of having the model work as an LM is the possibility of unsupervised model improvements, e.g. pre-training of the model's LM part from larger text corpora or 1-shot fine-tuning to new datasets with different word sequence patterns.

In order to address the problem of significantly imbalanced training data (the majority of the words in the corpus are fluent), we use a weighted cross-entropy loss in which the weight of a data point is inversely proportional to its label's frequency in the training set. Our overall loss function is of the form:
$$L = \mathit{WL}_{main} + \alpha L_{lm} + \frac{\lambda}{2}  \sum_{i} {w_i^2}$$

-- where $\mathit{WL}_{main}$ and $L_{lm}$ are respective losses for the disfluency tagging (class-weighted) and language modeling tasks (LM loss coefficient $\alpha$ is tuned empirically). The last term is L2 regularisation which we apply to the model's weight parameters $w_i$ (those of word embeddings, LSTM gates, and MLPs) leaving all the biases intact. L2 coefficient $\lambda$ is also tuned empirically (see Appendix \ref{ref:apx} for the values of the constants).

The model is implemented in Tensorflow \cite{tensorflow2015-whitepaper} and is openly available.

\section{Disfluency datasets and tags}\label{sec:data}


\subsection{The Switchboard dataset} For training our model, we use the Switchboard Dialog Acts dataset (SWDA) with manually annotated disfluency tags \cite{swda}. We use a pre-processed version of the dataset by \newcite{DBLP:conf/interspeech/HoughS15} containing 90,497 utterances with transformed tagging: following their convention, there are 27 tags in total consisting of: {\tt <f/>} tag for fluent tokens; {\tt <e/>} for edit tokens; {\tt <rm-\{n\}/>} tags for repair tokens that determine the start of the reparandum to be $n$ tokens/words back; and {\tt <rpSub>} \& {\tt <rpDel>} tags which mark the end of the \texttt{repair} and classify whether the repair is a \emph{substitution} or \emph{deletion} repair. The latter tokens can be combined with {\tt <rm-\{n\}>} tokens, which explains the total of 27 tags - see (\ref{repair-rnn}) for an example where the \texttt{repair} word, `Spanish', is tagged as \texttt{<rm-4><rpSub>} meaning this is a substitution repair that retraces $4$ tokens back from the current token. 

\begin{footnotesize}

\enumsentence{$\underset{\langle f/\rangle}{\textrm{with}} \underbrace{[\underset{\langle f/\rangle}{\textrm{Italian}}}_{reparandum}+\underbrace{\{\underset{\langle e/\rangle}{\textrm{uh}}\underset{\langle e/\rangle}{\textrm{no}}\underset{\langle e/\rangle}{\textrm{uh}}\}}_{interregnum} \underbrace{\underset{\langle rpSub\rangle}{\underset{\langle rm-4\rangle
}{\textrm{Spanish}]}}}_{repair} \ \ \underset{\langle f/\rangle}{\textrm{cuisine}}$}\label{repair-rnn}
\end{footnotesize}

The distribution of different types of tokens is highly imbalanced: only about 4\% of all tokens are involved in disfluency structures (the detailed statistics are shown in the Appendix \ref{ref:apx}). See above, Section \ref{model} for how our model deals with this.

\subsection{The bAbI+ dataset} To evaluate the cross data-set generalisation properties of our model and that of \newcite{DBLP:conf/interspeech/HoughS15}, we employ an additional dataset~-- bAbI+ introduced by \newcite{Shalyminov.etal17}. bAbI+ is an  extension of the original bAbI Task 1 dialogues \cite{babi} where different disfluency structures~-- such as hesitations, restarts, and corrections~-- can be mixed in probabilistically. Crucially these can be mixed in with complete control over the syntactic and semantic contexts in which the phenomena appear, and therefore the bAbI+ environment allows controlled, focused experimentation of the effect of different phenomena and their distributions on the performance of different models. Here, we use bAbI+ tools\footnote{See \url{https://bit.ly/babi_tools}} to generate new data for the controlled generalisation experiment\footnote{Data is available at \url{http://bit.ly/babi_plus_disfluencies_study}} of what kinds of disfluency phenomena are captured better by each model.

We focus here on the following disfluency patterns:
\begin{itemize}
    \item \textbf{Hesitations}, e.g.\ as in ``we will be \textit{uhm} eight'' (mixed in are single edit tokens);
    \item \textbf{Prepositional Phrase restarts (PP-restart)}, e.g. ``in a \textit{in a um in a} moderate price range'' (repair of a PP at its beginning with or without an interregnum);
    \item \textbf{Clausal restarts (CL-restart)}, e.g.\ ``can you make a restaurant \textit{uhm yeah can you make a restaurant} reservation for four people with french cuisine in a moderate price range'' (repair of the utterance from the beginning starting at arbitrary positions);
    \item \textbf{Corrections (NP and PP)}, e.g. ``with Italian \textit{sorry Spanish} cuisine'', as was initially discussed in Section \ref{sec:introduction}.
\end{itemize}

We generated independent bAbI+ datasets with each disfluency type. The disfluency phenomena above were chosen to resemble disfluency patterns in the original SWDA corpus (see Tables \ref{tab:common_repairs}, \ref{tab:pos_repair_patterns}, and \ref{tab:keyword_repair_patterns} for examples), as well as intuitive considerations for the phenomena relevant for goal-oriented dialogue (namely, corrections).

The intuition for a generalisation experiment with data like this is as follows: while having similar disfluency patterns, our bAbI+ utterances differ from SWDA in the vocabulary and the word sequences themselves as they are in the domain of goal-oriented human-computer dialogue~--- this property makes it possible to evaluate the generalisation capabilities of a model outside its training domain.

\section{Evaluation and experimental setup}\label{sec:eval}

\begin{table*}[t]
\begin{center}
\small
\begin{tabular}{|p{0.27\linewidth}|c|c|c|c|}\hline
\rowcolor{Gray}\textbf{Model}&\textbf{F\textsubscript{e}}&\textbf{F\textsubscript{rm}}&\textbf{F\textsubscript{rps}}\\\hline
\hline
\cite{DBLP:conf/interspeech/HoughS15}&0.902&0.711&0.689\\\hline
\cite{DBLP:conf/eacl/SchlangenH17}&0.918&---&0.719\\\hline
LSTM&0.915&0.693&0.775\\\hline
Multi-task LSTM&\textbf{0.919}&\textbf{0.753}&\textbf{0.816}\\\hline
\end{tabular}
\end{center} \caption{Evaluation of the disfluency tagging models}\label{tab:results}
\end{table*}

\begin{table*}[t]
\begin{center}
\small
\begin{tabular}{|p{0.25\linewidth}|c||c|c|c||c|c|c|}\hline
\rowcolor{Gray}\textbf{Model}&\textbf{hesitations (F\textsubscript{e} )}&\multicolumn{3}{c||}{\textbf{PP restarts}}&\multicolumn{3}{c|}{\textbf{CL-restarts}}\\
\rowcolor{Gray}&&\textbf{F\textsubscript{e}}&\textbf{F\textsubscript{rm}}&\textbf{F\textsubscript{rps}}&\textbf{F\textsubscript{e}}&\textbf{F\textsubscript{rm}}&\textbf{F\textsubscript{rps}}\\\hline
\hline
\cite{DBLP:conf/interspeech/HoughS15}&0.917&0.774&0.875&0.877&0.938&0.471&0.630\\\hline
LSTM&\textbf{0.956}&\textbf{1.0}&0.982&0.993&0.948&0.36&0.495\\\hline
Multi-task LSTM&0.910&\textbf{1.0}&\textbf{0.993}&\textbf{0.997}&\textbf{0.991}&\textbf{0.484}&\textbf{0.659}\\\hline
\end{tabular}
\end{center} \caption{Controlled generalisation evaluation}\label{tab:controlled_generalisation}
\end{table*}

We employ exactly the same evaluation criteria as \newcite{DBLP:conf/interspeech/HoughS15}: micro-averaged F1-scores for edit ($F_e$) and {\tt <rm-\{n\}/>} tokens ($F_{rm}$) as well as for whole repair structures ($F_{rps}$). We compare our Multi-task LSTM model to its single-task version (disfluency tag predictions only) as well as to the system of \newcite{DBLP:conf/interspeech/HoughS15} and the joint disfluency tagging/utterance segmentation model of \newcite{DBLP:conf/eacl/SchlangenH17} (all of the applicable word-level metrics on dialogue transcripts). These use a hand-crafted Markov Model for post-processing, whereas our model learns in an end-to-end fashion.

We train our model using the SGD optimiser and monitor the $F_{rm}$ on the dev set as a stopping criterion. The model's hyperparameters are tuned heuristically, the final values are listed in the Appendix \ref{ref:apx}. We use class weights in the main task's loss to deal with the highly imbalanced data, so that the weight of the $k^{th}$ class is calculated as $ W_k=1/{(C_k)^\gamma}$, where $C_k$ is the number of $k^{th}$ class instances in the training set, and $\gamma$ is a smoothing constant set empirically.

\subsection{Results}

The results are shown in Table \ref{tab:results}. Both single- and multi-task LSTM are able to outperform the \newcite{DBLP:conf/interspeech/HoughS15} model on edit tokens and repair structures, but the multi-task one performs significantly better on {\tt <rm-\{n\}/>} tags and surpasses both previous models. The reason $F_{rps}$ is higher than $F_{rm}$ in general is that due to the tag conversion, fluent tokens inside reparandums and repairs are treated as part of repair, and they contribute to the global positive and negative counters used in the micro-averaged F1.

Controlled generalisation experiment results are shown in Table \ref{tab:controlled_generalisation}~--- note that we could only run the model of \newcite{DBLP:conf/interspeech/HoughS15} on bAbI+ data because that of \newcite{DBLP:conf/eacl/SchlangenH17} works in a setup different from ours. 
It can be seen that the LSTM tagger is somewhat overfitted to edit tokens on SWDA. This is the reason it outperforms the Multi-task LSTM on the hesitations dataset and has a tied 1.0 on edit tokens on PP restarts dataset. In all other cases, Multi-task LSTM demonstrates superior generalisation.

As for NP/PP self-corrections which are not present in Table \ref{tab:controlled_generalisation}: none of the systems tested were able to handle these. Evaluation on the this dataset revealed $0.0$ accuracy with all systems. We discuss these results below. 

\begin{table}[t]
\small
\begin{minipage}{0.44\linewidth}
\begin{tabular}
{|p{0.15\linewidth}p{0.3\linewidth}c|}
\rowcolor{Gray}
\hline
\textbf{Repair length}&\textbf{Repair text}&\textbf{Frequency}\\\hline
\hline
1&i i \textit{i}&139\\
&the the \textit{the}&33\\
&and and \textit{and}&31\\
&it it \textit{it}&29\\
&its its \textit{its}&26\\\hline
2&it was \textit{it was}&67\\
&i dont \textit{i dont}&57\\
&i think \textit{i think}&44\\
&in the \textit{in the}&39\\
&do you \textit{do you}&23\\\hline
3&a lot of \textit{a lot of}&7\\
&that was \textit{uh that was}&5\\
&it was \textit{uh it was}&5\\
&what do you \textit{what do you}&4\\
&i i dont \textit{i dont}&4\\\hline
\end{tabular}

\caption{Most common repairs in SWDA}\label{tab:common_repairs}
\end{minipage}
\begin{minipage}{0.55\linewidth}
\begin{tabular}
{|p{0.25\linewidth}p{0.5\linewidth}p{0.1\linewidth}|}
\hline
\rowcolor{Gray}
\textbf{POS pattern}&\textbf{Examples}&\textbf{repairs \%}\\\hline\hline
DT NN DT NN&this woman this socialite&0.1\\
&a can a garage&\\
&the school that school&\\\hline
JJ NN JJ NN&high school high school&0.03\\
&good comedy good humor&\\
&israeli situation palestinian situation&\\\hline
DT UH DT NN&that uh that punishment&0.02\\
&the uh the cauliflower&\\
&that uh that adjustment&\\\hline
DT NN UH DD NN&a friend uh a friend&0.01\\
&a lot uh a lot&\\
&a lot um a lot&\\\hline
NN PRP VBP NN NN&ribbon you know hair ribbon&0.01\\
&thing you know motion detector&\\\hline
\end{tabular}
\caption{SWDA repairs by POS-tag pattern}\label{tab:pos_repair_patterns}
\end{minipage}
\end{table}

\begin{table}
\small
\centering
\begin{tabular}
{|lll|}
\hline
\rowcolor{Gray}
\textbf{Keyword pattern}&\textbf{Examples}&\textbf{repairs \%}\\\hline\hline
sorry\texttt{$<$e/$>$} *&or \textit{im sorry} no&0.02\\
&\textit{um im sorry} what&\\
&thank you \textit{im sorry} i just got home from work&\\\hline
sorry\texttt{$<$e/$>$} *\texttt{$<$rm-*/$>$}&and he told us theres two sixteen bit slots and two eight bit&0.009\\
&\textit{sorry two four sixteen bit slots and two eight bit} slots available for the user&\\\hline
i\texttt{$<$e/$>$} mean\texttt{$<$e/$>$} *&i mean&4\\
&i mean yeah&\\
&i mean uh&\\
&i mean i&\\\hline
i\texttt{$<$e/$>$} mean\texttt{$<$e/$>$} *\texttt{$<$rm-*/$>$}&i mean i i&0.5\\
&but i mean whats whats happened here is is is&\\
&i mean you youve&\\\hline
\end{tabular}
\caption{SWDA repairs by interregnum}\label{tab:keyword_repair_patterns}
\end{table}

\section{Discussion and future work}

We have presented a multi-task LSTM-based disfluency detection model which outperforms previous neural network-based incremental models while being significantly simpler than them.

For the first time, we have demonstrated the generalisation potential of a disfluency detection model by cross-dataset evaluation. As the results show, all models achieve reasonably high generalisation level on the very local disfluency patterns such as hesitations and PP restarts. However, the accuracy drops significantly on less restricted restarts spanning arbitrary regions of utterances from the beginning. On the majority of those disfluency patterns, our model achieves a superior generalisation level.

Interestingly, none of the models were able to detect NP or PP corrections such as those often glossed in disfluency papers (e.g.\ ``A flight to Boston uh I mean to Denver''). The most likely explanation for this could be the extreme sparsity of such disfluencies in the SWDA dataset. 

We performed analysis of SWDA disfluencies in order to explore this hypothesis and examined their distribution based on length in tokens and POS-tag sequence patterns of interest. As shown in Tables \ref{tab:common_repairs} and \ref{tab:pos_repair_patterns}, the vast majority of disfluencies found are just repetitions without speakers actually correcting themselves. This observation is in line with prior studies, showing that the distribution of repair types varies significantly across domains \cite{Colman.Healey11}, modalities \cite{Oviatt95}, and gender \& age groups \cite{Bortfeld.etal01}~--- see \newcite{Purver.etal18} for a nice discussion.

While this is very likely the correct explanation, we cannot rule out the possibility that such self-corrections are inherently more difficult to process for particular models - that needs a separate experiment that holds frequency of particular repair structures constant in the training data. 


Addressing this issue is our next step since we designed the multi-task LSTM with this in mind. As such, we will explore possibilities of knowledge transfer to new closed domains in a 1-shot setting, both with regular supervised training and unsupervised LM fine-tuning.

\section{Acknowledgements}
We are very grateful to Julian Hough for offerring extensive help with running their 2015 disfluency detection model \cite{DBLP:conf/interspeech/HoughS15} and their SWDA tagging conventions, as well as constructive feedback throughout.

\bibliography{all,semdial2018}
\bibliographystyle{acl}

\begin{appendices}
\section{} \label{ref:apx}

\begin{table}[h]
\centering
\small
\begin{tabular}{|p{0.3\linewidth}|p{0.3\linewidth}|}
\hline
\rowcolor{Gray}
\textbf{Parameter}&\textbf{Value}\\\hline
\hline
optimiser&stochastic gradient descent\\\hline
loss function&weighted cross-entropy\\\hline
vocabulary size&6157\\\hline
embedding size&128\\\hline
MLP layer sizes&[128]\\\hline
learning rate&0.01\\\hline
learning rate decay&0.9\\\hline
batch size&32\\\hline
$\alpha$&0.1\\\hline
$\lambda$&0.001\\\hline
$\gamma$&1.05\\\hline
\end{tabular}
\caption{Multi-task LSTM training setup}\label{tab:lstm_setup}
\end{table}

\begin{table}[h]
\small
\centering
\begin{tabular}
{|p{0.2\linewidth}|p{0.28\linewidth}|p{0.1\linewidth}|}
\hline
\rowcolor{Gray}
\textbf{Label type}&\textbf{Label}&\textbf{Frequency}\\\hline
\hline
fluent token&\texttt{<f/>}&574771\\\hline
edit token&\texttt{<e/>}&45729\\\hline
single-token substitution&\texttt{<rm-\textit{\{1-8\}}/><rpEndSub/>}&13003\\\hline
single-token deletion&\texttt{<rm-\textit{\{1-8\}}/><rpEndDel/>}&1011\\\hline
multi-token substitution start&\texttt{<rm-\textit{\{1-8\}}/><rpMid/>}&6976\\\hline
multi-token substitution end&\texttt{<rpEndSub>}&6818\\\hline
\end{tabular}
\caption{SWDA labels}\label{tab:swda_labels}
\end{table}

\end{appendices}

\end{document}